\documentclass{ws-us}

\usepackage{algorithm}
\usepackage{algorithmicx}
\usepackage[algo2e,ruled]{algorithm2e} %带竖线

\usepackage{amssymb,epsfig,color,amsmath,amsfonts,mathrsfs}
\usepackage{epstopdf}
\usepackage{epsfig}
\usepackage{lipsum}
\usepackage{amsfonts}
\usepackage{graphicx}
\usepackage{epstopdf}

\usepackage{amssymb,bm}
\usepackage{multirow}
\usepackage{hyperref}

\usepackage{cite}
\usepackage{amsmath,amssymb}
\usepackage{subfigure}
\usepackage{algpseudocode}
\usepackage[T1]{fontenc}
\usepackage{float}
\usepackage{adjustbox}
\usepackage{setspace}
\usepackage{booktabs}
\usepackage{multirow}

\begin{document}
\setlength{\pdfpagewidth}{8.5in}
\setlength{\pdfpageheight}{11in}
%\catchline{0}{0}{2013}{}{}

%\markboth{Zhaohan Feng, et al's paper}{Learning Hybrid Policies for MPC with Application to Drone Flight in Unknown Dynamic Environments}

\title{Learning Hybrid Policies for MPC with Application to Drone Flight in Unknown Dynamic Environments}

\author{
\let\thefootnote\relax\footnotetext{Email address: $^*$gangwang@bit.edu.cn}
\let\thefootnote\relax\footnotetext{The work was supported in part by the National Natural Science Foundation of China under Grants 62173034, U23B2059, 61925303, 62088101, and  the Natural Science Foundation of Chongqing under Grant 2021ZX4100027. %the National Key R\&D Program of China under Grant 2021YFB1714800,  
}
Zhaohan Feng$^{\dag\ddag}$, Jie Chen$^{\dag\S}$, Wei Xiao$^{\dag\ddag}$, Jian Sun$^{\dag\ddag}$, Bin Xin$^{\dag}$, Gang Wang$^{\dag\ddag}$}
\address{$^{\dag}$National Key Lab of Autonomous Intelligent Unmanned Systems, Beijing Institute of Technology, Beijing, China\\
$^{\ddag}$Beijing Institute of Technology Chongqing Innovation Center, Chongqing, China\\
$^{\S}$Department of Control Science and Engineering, Tongji University, Shanghai 201804, China
}

\maketitle

\allowdisplaybreaks

\begin{abstract}
In recent years, drones have found increased applications in a wide array of real-world tasks. Model predictive control (MPC) has emerged as a practical method for drone flight control, owing to its robustness against modeling errors/uncertainties and external disturbances. However, MPC's sensitivity to manually tuned parameters can lead to rapid performance degradation when faced with unknown environmental dynamics. This paper addresses the challenge of controlling a drone as it traverses a swinging gate characterized by unknown dynamics. This paper introduces a parameterized MPC approach named hyMPC that leverages high-level decision variables to adapt to uncertain environmental conditions. To derive these decision variables, a novel policy search framework aimed at training a high-level Gaussian policy is presented. Subsequently, we harness the power of neural network policies, trained on data gathered through the repeated execution of the Gaussian policy, to provide real-time decision variables. The effectiveness of hyMPC is validated through numerical simulations, achieving a 100\% success rate in 20 drone flight tests traversing a swinging gate, demonstrating its capability to achieve safe and precise flight with limited prior knowledge of environmental dynamics. 
\end{abstract}


\begin{multicols}{2}
\section{Introduction}
In recent years, drones have been increasingly deployed for diverse real-world tasks. Leveraging state-of-the-art model-based approaches, drones have demonstrated remarkable agility in executing maneuvers within both static and dynamic environments \cite{science2021Scaramuzza,eng2022chen,zhang2023review, Zhang2023UAVPlanning}. This rapid technological advancement has given rise to numerous autonomous drone racing series and challenges \cite{droneracing1,droneracing4,dronechallenge}. These endeavors necessitate drones to autonomously and optimally navigate static or dynamic environments solely relying on visual information. An illustrative scenario is depicted in Fig. \ref{fig:alphapilot}, where a drone embarks from an initial podium and adeptly traverses a sequence of static gates to reach a predefined target location \cite{dronechallenge}. However, the computational constraints posed by on-board computing resources often present substantial challenges for drones in accomplishing these tasks.

Drones encounter a wide spectrum of technical challenges, spanning the domains of stable flight control, robust gate detection and precise registration during trajectory planning, accurate tracking control, and the detection and prediction of dynamic obstacles \cite{droneracing1,zhou2023efficient,song2023science,qin2023time}. Addressing this formidable array of challenges has been notably expedited through the integration of advanced sensing and perception algorithms. In the context of autonomous drone racing, many of these advancements empower drones to expertly navigate environments featuring static gates and obstacles. Some research also considers the challenge of moving gates with a limited range of motion and relatively slow speed \cite{droneracing4,dronechallenge}. 

Model predictive control (MPC) \cite{mpcbook} stands as a widely embraced model-based approach for drone flight control, lauded for its proficiency in delivering high-accuracy solutions. This proficiency arises from the real-time resolution of constrained optimization problems and the inherent robustness it exhibits against modeling errors and external disturbances \cite{mpcondrone3,unmansyst3,Liu2023TACMPC1,Liu2023TACMPC2}. In the pursuit of mission success, drones must exhibit the capability to instantaneously perceive environmental changes and dynamically recalibrate their trajectories in response.
However, the performance of MPC is intricately interwoven with several manually designed components, notably the cost function, prediction horizon, and weight parameters. Achieving a satisfactorily refined solution often necessitates a labor-intensive process of hyper-parameter tuning, reliant on trial-and-error methodologies. Consequently, in dynamic environments mandating real-time parameter adaptations and constrained by prediction horizons imposed to meet exacting control frequency requirements, the performance of MPC controllers can exhibit significant susceptibility to degradation.

\begin{figurehere}
	\begin{center}
		\centerline{\includegraphics[width=3in]{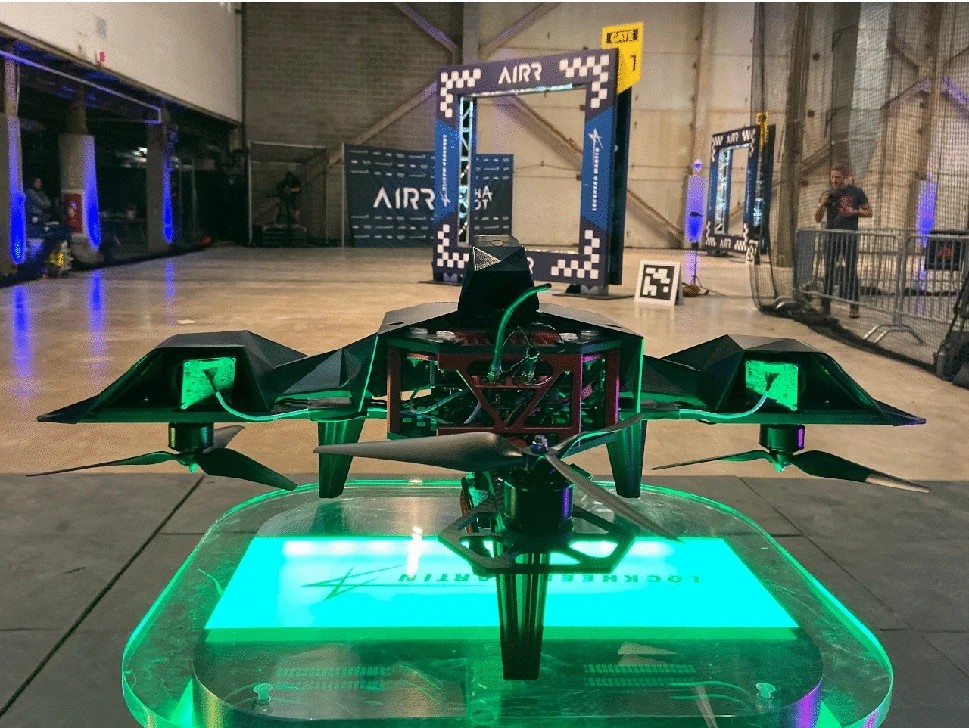}}
		\caption{AlphaPilot: A drone departing from a start point and autonomously navigate through some gates by vision\cite{dronechallenge}.}
		\label{fig:alphapilot}
	\end{center}
\end{figurehere}

Reinforcement learning (RL) offers a principled approach to designing control policies within a spectrum of complex environments \cite{rlbook,jasrl, zhang2023storm}, by directly learning from data \cite{Wang2022SCICHNhybrid}. RL methods can effectively manage high-dimensional observations when coupled with deep neural networks, while they frequently grapple with challenges related to data inefficiency and suboptimal generalization performance, often lacking the robustness of stability guarantees.
To harness the strengths of real-time MPC and offline RL, thereby amalgamating MPC's prowess in high-quality trajectory optimization with RL's near-optimal decision-making capabilities, becomes an enticing prospect. Taking the exemplar case of AlphaPilot, navigating swinging gates of known dynamics, \cite{highmpc2} introduced high-MPC, a framework harmoniously uniting RL and MPC. In this framework, high-level decision variables concerning gate traversal times are derived via policy search, while the lower-level parameterized trajectory planning is entrusted to MPC.
In essence, the task revolves around skillfully traversing gates at their anticipated positions and precisely timed intervals. When armed with knowledge of gate dynamics, the prediction of traversal times and associated points becomes relatively straightforward. However, in scenarios marked by the ambiguity of unknown gate dynamics, the task escalates in complexity.

In this work, we extend upon the foundations established in high-MPC \cite{highmpc2} and significantly broaden its scope to encompass the autonomous flight of drones through dynamically moving gates characterized by \emph{unknown} dynamics. To achieve this objective, a novel policy search algorithm is introduced geared toward learning high-level decision variables, which guide MPC in addressing the inherent challenges posed by the \emph{unknown} dynamics.
By harnessing a neural network framework to predict gate dynamics, the gate-traversing flight task is divided into two distinct subtasks: tracking the gate's motion and executing flight trajectories that align with the predicted gate position and timing. These two subtasks are methodically addressed by incorporating meticulously designed cost functions within the MPC framework. The high-level policy is trained to optimize the mix of these associated cost functions, thus enabling MPC to seamlessly accommodate both objectives in a near-optimal manner.
To continually enhance the policy's performance, we employ a weighted maximum likelihood approach \cite{policysearch} for iterative policy parameter updates. Furthermore, by leveraging a self-supervised training method as proposed in \cite{highmpc2}, neural network policies were trained to provide real-time decision variables, grounded in the drone's observations, alongside predictions of gate dynamics.
Through comprehensive numerical simulations, we empirically validate the effectiveness of the proposed algorithms. Employing this innovative methodology, drones exhibit enhanced precision when traversing swinging gates marked by \emph{unknown} dynamics, all while operating within the constraints of a limited MPC prediction horizon.

\section{Preliminaries and System Modeling}

\subsection{Drone dynamics}
A drone's state is commonly described from the inertial frame $I$ to the drone's body frame $B$, as $\bm{s}=[\bm{p}_{IB},\bm{v}_{IB},\bm{q}_{IB},\bm{\omega}_B]^T$ where $\bm{p}_{IB}\in\mathbb{R}^3$ is the position, $\bm{v}_{IB}\in\mathbb{R}^3$ is the velocity, $\bm{q}_{IB}\in\mathbb{SO}(3)$ is the unit quaternion which describes the pose of the drone's platform and $\bm{\omega}_B=[\omega_x, \omega_y,\omega_z]^T$ is the body rate of the drone \cite{RAL2022Romero}. We denote $\bm{f}_w=[0,0,c]^T$ as the mass-normalized thrust acting on the drone. The dynamic equations are proposed as follows
\begin{equation}\label{eq:quaddynamics}
    \begin{aligned}
\dot{\bm{p}}_{IB}&=\bm{v}_{IB}\\
    \dot{\bm{v}}_{IB}&=\bm{q}_{IB}\odot\bm{f}_{w}-\bm{g}\\
    \dot{\bm{q}}_{IB}&=\frac{1}{2}\bm{\Lambda}(\bm{\omega}_B)\cdot\bm{q}_{IB}\\
    \dot{\bm{\omega}}_B&=\bm{J}^{-1}(\bm{\eta}-\bm{\omega}_B\times\bm{J}\bm{\omega}_B)
\end{aligned}    
\end{equation}
where $\bm{\Lambda}$ is a skew-symmetric matrix, $\bm{J}$ is the drone's inertia matrix and $\bm{\eta}\in\mathbb{R}^3$ is the torque acting on the drone. For convenience, we select $\bm{x}=[ \bm{p}_{IB},\bm{v}_{IB},\bm{q}_{IB} ]^T$ as the drone's state and $\bm{u}=[c,\omega_x,\omega_y,\omega_z]^T$ as the drone's control input during trajectory planning. 

Generally, the dynamics governing a drone's behavior can be succinctly expressed through a set of nonlinear ordinary differential equations, denoted as $\dot{\bm{x}}=f(\bm{x}, \bm{u})$. In this equation, $\bm{x}\in\mathbb{R}^{10}$ represents the system state vector, while $\bm{u}\in\mathbb{R}^4$ represents the control input vector. 

\subsection{Model predictive control}

In practical application, the continuous-time model characterizing a drone's behavior undergoes discretization and is approximated as $\bm{x}_{h+1} = \bm{x}_h + d\tilde{f}(\bm{x}_h,\bm{u}_h)$, where $d$ signifies the discrete time interval, $\tilde{f}$ represents the approximated dynamical model, and $h$ denotes the discrete time step. At each time instance $t$, a cost function $c_t = c(\bm{x}_t, \bm{u}_t; \bm{\textbf{p}}, \bm{\textbf{t}}_p, \lambda)$ is formulated. Here, $\bm{\textbf{p}}$ assembles a series of reference system states, $\bm{\textbf{t}}_p$ corresponds to the expected time for the system to attain these reference states, and $\lambda$ represents a high-level decision variable. In the context of this study, $\lambda$ is defined as the weighting factor dictating the mixture of cost functions associated with subtasks. 
Hence, MPC employs the current state ${\bm x}_t$ as its starting point ${\bm x}_{t_0}$, and reference states $\bm{\textbf{p}}$ along with expected reaching time $\bm{\textbf{t}}_p$, and generates a predicted trajectory $\varGamma^* = \{\bm{u}_{t_0}^*, \bm{x}_{t_1}^*, \bm{u}_{t_1}^*,...,\bm{u}_{t_{H-1}}^*, \bm{x}_{t_H}^*\}$ with time interval $d=t_{\tau+1}-t_\tau$ by minimizing a manually selected cost function $C=\sum_{H}c_{t_h}$ over a finite time horizon $H$ as follows
\begin{subequations}	 \label{eq:mpcproblem}
	\begin{align}
		\min_{\bm{u}_{t_0:t_{H-1}}\atop \bm{x}_{t_1:t_{H}}}~&~	C=\sum_{t=t_0}^{t_{H}}c(\bm{x}_t,\bm{u}_t; \bm{\textbf{p}}, \bm{\textbf{t}}_p, \lambda)+c(\bm{x}_{t_{H}}; \bm{\textbf{p}}, \bm{\textbf{t}}_p, \lambda) \\
		{\rm s.t.} 
  \quad~&~-\bm{\omega}_{max}\leq \bm{\omega}_{B}\leq \bm{\omega}_{max} \\
				&~\bm{x}_{t_{\tau+1}}=\bm{x}_{t_\tau}+d\tilde{f}(\bm{x}_{t_\tau},\bm{u}_{t_\tau})
				\label{eq:sub1c}
	\end{align}
	\end{subequations}
 where $-\bm{\omega}_{max}\leq \bm{\omega}_{B}\leq \bm{\omega}_{max}$ represents the bodyrate limit of the drone and $\bm{x}_{t_{\tau+1}}=\bm{x}_{t_\tau}+d\tilde{f}(\bm{x}_{t_\tau},\bm{u}_{t_\tau})$ is the dynamic constraint of the drone defined by its dynamical model. 
 
 After generating $\varGamma^*$, the next control input to be executed is selected as $\bm{u}=\bm{u}_{t_0}^*$.  Typical drone flight tasks,  e.g., AlphaPilot, usually have three elements specified: initial state $\bm{x}_{t_0}$, intermediate state (a.k.a. waypoint) $\bm{x}_{\rm way}$ with an expected passing time $t_{\rm way}$, and target state $\bm{x}_{\rm target}$. By running MPC at each time $t$ with inputs $\bm{x}_{t_0}=\bm{x}_t$, $\bm{\textbf{p}}:=\{\bm{x}_{\rm way},\bm{x}_{\rm target}\}$ and $\bm{\textbf{t}}_p:=\{t_{\rm way}\}$, MPC is capable of generating a trajectory starting from $\bm{x}_{t_0}$, passing $\bm{x}_{\rm way}$ at time $t_{\rm way}$ and after that, trying to get as close to $\bm{x}_{\rm target}$ as possible. 

 \subsection{Policy search}
Policy search stands as an effective method in the realm of robot control \cite{rlbook}. Its primary objective is the discovery of an optimal policy by exploring policy parameters and maximizing the expected returns derived from a series of rollout trajectories generated by the policy. The fundamental aim of policy search lies in the optimization of the expected return associated with the policy-generated trajectories, expressed as
 \begin{align}
    J_{\bm{\theta}}=\mathbb{E}_{p(\bm{\tau};\bm{\theta})}[R(\bm{\tau})]=\int p(\bm{\tau};\bm{\theta})R(\bm{\tau})d\bm{\tau}
 \end{align}
 where $\bm{\theta}$ is the sampled policy parameters, $R(\bm{\tau})$ is the total return of a single sampled trajectory and $p(\bm{\tau};\bm{\theta})$ is the probability of the trajectory under $\bm{\theta}$. Specifically, the trajectory distribution $p(\bm{\tau};\bm{\theta})$ is calculated using the formula
 \begin{align}
    p(\bm{\tau};\bm{\theta})=p(\bm{s}_1)\prod \limits_{t=1}^{T}\pi(\bm{a}_{t}|\bm{s}_{t};\bm{\theta})p(\bm{s}_{t+1}|\bm{s}_{t}, \bm{a}_{t})
 \end{align}
 where $\bm{s}_1$ is the initial state of the trajectory, $T$ is the total time horizon of the trajectory, $\pi(\bm{a}_{t}|\bm{s}_{t};\bm{\theta})$ is the action distribution under parameters $\bm{\theta}$ and $p(\bm{s}_{t+1}|\bm{s}_{t}, \bm{a}_{t})$ stands for the state-transition distribution. 
 
Policy search methodologies fall into two distinct categories: step-based policy search and episode-based policy search \cite{policysearchbook}. In step-based policy search, perturbations are introduced into the policy's output space at each decision-making step, leading to the generation of stochastically guided rollout trajectories \cite{dpg,2ndpg}. The reward-to-come of states in these rollout trajectories is employed to calculate the expected value over the trajectory distribution. Step-based policy search is commonly employed when training end-to-end policies that directly generate control commands from state observations \cite{Jemin2019science}. However, the intricacies of robot dynamical models often result in a lack of smoothness in end-to-end learned control policies, thus diminishing their practicality.
 
Episode-based policy search is often deemed a more practical approach for implementation in the realm of robotics. In episode-based policy search, perturbations are introduced into the policy's parameter space at the onset of each episode \cite{Peters_Mulling_Altun_2010,DBLP,tdmpc}. These perturbations yield variations in policy parameter sampling, consequently resulting in distinct rollout trajectories. The policy parameters are subsequently updated utilizing the total reward attained from sampled trajectories. Episode-based policy search methods facilitate the acquisition of parameterized motor primitives and intricate skills that are challenging to engineer artificially. These acquired primitives and skills prove highly effective in tandem with lower-level actuators such as MPC, PID controllers, and more. The utilization of episode-based policy search methods empowers the development of safer and more sophisticated control policies, thereby rendering them eminently practical for real-world robotics applications. 

\section{RL-driven MPC Trajectory Optimization}

In this work, we address the challenge of maneuvering drones through a swinging gate characterized by \emph{unknown} dynamics. To tackle this task, an RL-driven MPC framework is introduced. Given the assumption of unknown gate dynamics, a multi-layer perceptron (MLP) denoted as $f_w$ is employed to acquire a dynamical model. The MLP takes historical gate states $\bm{x}_{t-d}^g$ and $\bm{x}_{t}^g$ as inputs and predicts the gate's future state $\bm{x}_{t+d}^g$, where $d$ signifies the time interval. The parameters of $f_w$ are updated online through back-propagation of the prediction error $\|\bm{x}_{t+d}^g-\overline{\bm{x}}_{t+d}^g\|^2$ at time $t+d$. Here, $\bm{x}_{t+d}^g$ and $\overline{\bm{x}}_{t+d}^g$ denote the actual gate state and its corresponding prediction at time $t+d$, respectively. By leveraging the learned $f_w$, we can promptly forecast the gate's motion.

\begin{figurehere}
	\begin{center}
		\centerline{\includegraphics[width=3.4in]{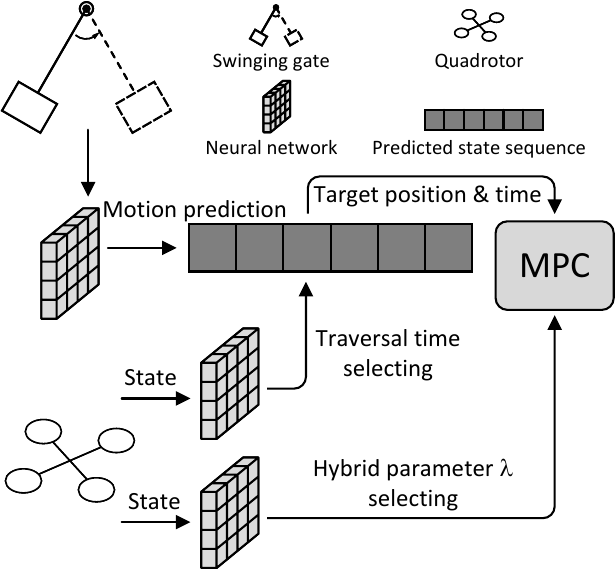}}
		\caption{A block diagram about the proposed hyMPC dealing with the task of traversing a swinging gate. }
		\label{fig:blockdiagram}
	\end{center}
\end{figurehere}

In our RL-driven MPC framework, MPC serves as a parameterized low-level controller, receiving inputs including the current state $\bm{x}_{t_0}$, reference points $\bm{\textbf{p}}=\{\bm{x}_{\rm way}, \bm{x}_{\rm target}\}$, expected reaching times $\bm{\textbf{t}}_p=\{t_{\rm way},t_{\rm target}\}$, and a high-level decision variable $\lambda$.

To address motion prediction errors, we divide the traversal task into two distinct subtasks. Within the MPC framework, these subtasks are explicitly defined through separate cost functions. The first subtask involves tracking the gate's motion, achieved by setting $\bm{x}_{\rm target}$ to the current state of the gate, denoted as $\bm{x}_t^g$. The cost function associated with this subtask, governing the act of following the gate, is represented as

\begin{align}
c_{\rm follow}\!=\sum_{t=t_0}^{t_H}\left[L_{u_t}+\xi_{ f}(\bm{x}_t^q-\bm{x}_{t_0}^g)^TQ_{ f}(\bm{x}_t^q-\bm{x}_{t_0}^g)\right]
\label{eq:cfollow}
\end{align}
where $L_{u_t}=\bm{u}_{t}^{T}Q_{u}\bm{u}_t$ is a regularization cost for the control input, $\xi_{ f}=\eta{\exp}(-(t-t_{ f})^2)$ is an exponentially decaying weight parameter with $\eta>0$ defining the temporal spread of the weights for the gate-following subtask, $\bm{x}_t^q$ specifies the drone's state, $t_{f}$ represents the expected time at which the drone achieves the followed state $\bm{x}_t^g$, and $Q_{u}, Q_{f}$ are positive diagonal weight matrices. 

The other subtask is to traverse the gate at predicted state and time and reach the target state $\bm{x}_{\rm target}$. Let $\overline{\bm{x}}_{t_{p}}^g$ stand for the predicted state of the gate at  time $t_{p}$, where $t_{p}$ denotes the predicted passing time, generated by the high-level RL algorithm. Setting $\bm{x}_{\rm way}=\overline{\bm{x}}_{t_{p}}^g$ and $\bm{\textbf{t}}_p=\{t_p\}$, the cost function for passing the gate is defined as follows
\begin{align}
c_{\rm pass}
&=\sum_{t=t_0}^{t_{H-1}}\left[L_{{u}_t}+\xi_{p}(\bm{x}_t^q-\overline{\bm{x}}_{t_{p}}^g)^TQ_{ p}(\bm{x}_t^q-\overline{\bm{x}}_{t_{p}}^g)\right]\nonumber\\
&\quad +(\bm{x}_{t_H}^q-\bm{x}_{\rm target})^TQ_{ g}(\bm{x}_{t_H}^q-\bm{x}_{\rm target})
\label{eq:cpass}
\end{align}
where $\xi_{p}=\eta\exp(-(t-t_{p})^2)$ is an exponentially decaying weight, and $Q_{ p}$, $Q_{g}$ are positive diagonal weight matrices.

To ensure the generation of a successful gate-traversing flight, we employ a hybrid MPC cost function for real-time trajectory optimization. This function is meticulously crafted by skillfully combining the subtasks of gate-following and gate-traversing
\begin{equation}
C={\lambda}c_{\rm follow}+(1-\lambda)c_{\rm pass}
\label{eq:cfollow+cpass}
\end{equation}
where $\lambda\in [0,1]$ is a learnable hyper-parameter balancing the two subtasks. The MPC objective shifts to gate-following when $\lambda$ approaches $0$; otherwise, it focuses on gate traversal at the predicted state $\overline{\bm{x}}_{t_{p}}^g$ at time $t_{p}$ when $\lambda$ approaches $1$. 

\section{Episode-based Policy Search}
\subsection{Policy update}
To address the prediction error $\|\bm{x}_{t_{p}}^g-\overline{\bm{x}}_{t_{p}}^g\|$ associated with gate dynamics in real-time, we employ a parameterized policy $\lambda\sim \pi_\theta$ to search for the optimal high-level decision variable $\lambda$.  
To achieve this, we employ an episode-based policy search method, also known as episodic RL, as proposed in \cite{highmpc2}. The episode-based policy search introduces exploration noise in the policy parameter space at the start of each episode. Consequently, a series of diverse decision variables $\lambda$ are sampled from $\pi_\theta$. The quality of a trajectory $\varGamma$ generated by MPC with sampled $\lambda$ is evaluated using a specified reward function $R(\varGamma)$. 
The objective of finding the optimal decision variable $\lambda$ that maximizes the average return of sampled trajectories $R(\varGamma)$ is transformed into the optimization of the policy $\pi_\theta$. This is achieved by updating the parameters $\theta$ as follows

\begin{equation}
\max_{\theta}  \mathbb{E} \left[R(\varGamma)|\theta\right]\approx\log\int_\varGamma{R(\varGamma)p_\theta(\varGamma)d\varGamma}
\label{eq:policymaximize}
\end{equation}
where $p_{\theta}(\varGamma)$ represents the probability of generating trajectory $\varGamma$ while taking the policy parameters $\theta$. In practical scenarios, however, evaluating the integral in \eqref{eq:policymaximize} becomes computationally intractable. Following the approach presented in \cite{highmpc2}, one can introduce a variational distribution $q(\varGamma)$ to decompose the marginal log-likelihood in \eqref{eq:policymaximize}. Subsequently, the Monte-Carlo expectation-maximization (MC-EM) algorithm can be employed to approximate the solution of \eqref{eq:policymaximize} as follows
\begin{equation}
\theta^*:=\arg\max_\theta\sum_i\exp\{\zeta R(\varGamma^i)\}\log p_{\theta}(\varGamma^i)
\label{eq:mcemupdate}
\end{equation}
where $\zeta \geq 0$ denotes the inverse temperature of the soft-max distribution, and $\varGamma^i$ represents the $i$-th trajectory generated by MPC associated with $\lambda^i$ sampled from $\pi_{\theta}$. The softmax distribution encourages the policy to prioritize trajectories with higher expected rewards. By tuning the parameter $\zeta$, the policy update process can become more or less conservative. 

\vspace{0.3cm}
\begin{algorithm2e}[H]
	\SetAlgoNoLine %去掉竖线
	\caption{\textbf{Learning high-level Gaussian policies}}
	\label{alg:1}
	\KwIn{$\pi_\theta(\mu, \sigma^2)$,  $N$, MPC, $\bm{x}_{t_0}^q$, $\bm{x}_{\rm target}$, $f_w$}
	\While{not converged}{
		\setlength{\parindent}{1em}
		sample $N$ $\lambda$'s from $\pi_\theta(\mu, \sigma^2)$: $[\lambda]=[\lambda^1,\lambda^2,...,\lambda^N]$\\
		$[R]_\lambda=[ ]$\\
		\For{$\lambda^i$ in $[\lambda]$}{
			\setlength{\parindent}{2em}
			$[R]_i=[]$\\
			\For{$t_j$ in the trustworthy prediction horizon $\mathcal{T}_p$}{
				\setlength{\parindent}{3em}
				$\overline{\bm{x}}_{t_j}^g={f_w}(\bm{x}_{t_0-d}^g, \bm{x}_{t_0}^g, t_j)$\\
				$\varGamma_{j}^{i}={\rm MPC.solve}(\bm{x}_{t_0}^q,\bm{x}_{t_0}^g,\overline{\bm{x}}_{t_j}^g,\bm{x}_{\rm target},\lambda^i)$\\
				$[R]_i{\rm .append}(R(\varGamma_{j}^{i}))$\\
			}
			$R_\lambda^i=\mathop{\rm max}_j[R]_i,t_\lambda^i=t_j$\\
			$[R]_\lambda{\rm .append}(R_\lambda^i)$\\
		}
		$\mu^*,\sigma^*=\arg\max_{\mu,\sigma}\sum_i\exp(\zeta R_\lambda^i)\log p_\theta(\lambda^i)$\\
	}
	Output: Learned Gaussian policy $\pi_{\theta^*}(\mu^*,\sigma^{*2})$
\end{algorithm2e}
\vspace{0.3cm}

To seek the optimal high-level policy for trajectory planning, we meticulously craft the reward function $R(\varGamma)$ as follows. 
Considering the limited finite prediction horizon of MPC, we classify trajectories into three categories: trajectories that have not yet crossed the gate (not reached), trajectories that have crossed the gate's plane but failed to traverse it (failure), and trajectories that have crossed the gate's plane and successfully traversed it (success). For trajectories that have crossed the gate's plane, there must exist at least one pair of neighboring waypoints $\bm{x}_{t_{i-1}}^q$ and $\bm{x}_{t_i}^q$ within the trajectory, positioned on opposite sides of the gate's plane. In other words, $\bm{x}_{t_{i-1}}^q$ and $\bm{x}_{t_i}^q$ must satisfy the following constraint
\begin{equation}
\big[\big(\bm{x}_{t_{i-1}}^g-\bm{x}_{t_{i-1}}^q\big)\cdot\bm{\alpha}\big]\big[(\bm{x}_{t_{i-1}}^g-\bm{x}_{t_i}^q)\cdot\bm{\alpha}\big]\leq 0
\label{eq:trajclassification}
\end{equation}
where $\bm{\alpha}$ represents the normal vector of the gate's plane, and the $\cdot$ symbol denotes the dot product. For trajectories that have crossed the gate's plane, the traversal error at time $t_{p}$, denoted as $\Vert \bm{x}_{t_i}^q-\bm{x}_{t_i}^g\Vert$, serves as a criterion to determine whether the trajectory successfully traversed the gate or not. Specifically, if $\Vert \bm{x}_{t_i}^q-\bm{x}_{t_i}^g\Vert\geq\varepsilon$ for a given accuracy threshold $\epsilon>0$, the trajectory is labeled as a failure due to the traversal error being relatively too large. Conversely, if the traversal error is below this threshold, the trajectory is considered a success. The reward $R(\varGamma)$ assigned to trajectories that have crossed the gate's plane is calculated as follows
	\begin{equation}
		R(\varGamma)=
		\begin{cases}
			-\sum_{t=t_1}^{t_H}\psi_t\|\bm{x}_t^q-\bm{x}_t^g\|^2-t_{p},&{\rm success}\\
			-\Omega-\sum_{t=t_1}^{t_H}\psi_t\|\bm{x}_t^q-\bm{x}_t^g\|^2-t_{p},&{\rm failure}
		\end{cases}		\label{eq:rewardofarrive}
	\end{equation}
where $		\psi_t=\exp(-\varrho(t-t_{p})^2)$ for all $\varrho\in\mathbb{R}_+\nonumber$,
$\Omega\ge 0$ is a large constant for penalizing failed trajectories, $\psi_t$ denotes the exponential weight of gate-following, and $\varrho$ defines the temporal spread of $\psi_t$ similarly to \eqref{eq:cfollow}. 

To emphasize trajectories with shorter traversal times, the reward function in Eq. \ref{eq:rewardofarrive} is penalized by $t_p$. If there are no $\bm{x}_{t_{i-1}}^q$ and $\bm{x}_{t_i}^q$ satisfying  \eqref{eq:trajclassification}, then the corresponding trajectory is considered as not having yet crossed the gate. The reward $R(\varGamma)$ for such trajectories is computed as follows
\begin{equation}
\label{eq:rewardofnotarrive}
R(\varGamma)=-\sum_{t=t_1}^{t_H}\psi_t\|\bm{x}_t^q-\bm{x}_t^g\|^2
\end{equation}
where we have defined the coefficient $\psi_t=\exp(-\varrho(t-t_H)^2)$ for all $\varrho\in\mathbb{R}_+$.

For the present problem, we opt to learn a Gaussian policy denoted as $\pi_\theta(\lambda|\mu, \sigma^2)$, where both the mean and variance are considered as unknown variables. The pseudo-code detailing the proposed episode-based policy search algorithm for optimizing $\pi_\theta$ is presented in Algorithm~\ref{alg:1}. 

As the accuracy of gate motion prediction decreases over time, the predicted traversal time remains meaningful and reliable only within a reasonably limited time horizon. Therefore, for practical implementation, we introduce the concept of a trustworthy prediction horizon, denoted as $\mathcal{T}_p=\{t_0,t_1,...,t_{H_p}\}$. Instead of directly predicting the traversal time as done by a Gaussian policy in \cite{highmpc2}, employing a trustworthy prediction horizon allows us to consider all times $t_j\in \mathcal{T}_p$ as potential gate-traversing times and select the one that yields the highest trajectory reward.
At each time $t$ during the traversal, we initialize the input $\bm{x}_{t_0}^q$ with the current drone state $\bm{x}_{t}^q$ and reset the policy $\pi_\theta(\mu, \sigma^2)$ according to Algorithm~\ref{alg:1}. At the start of each iteration, we draw a set of $N$ values of $\lambda$ from the initialized policy $\pi_{\theta}$. For each realization, we compute its corresponding reward $R(\varGamma_{j}^{i})$ using equations \eqref{eq:rewardofarrive} and \eqref{eq:rewardofnotarrive}, where $\varGamma_{j}^{i}$ represents the trajectories generated by MPC. The solution to equation \eqref{eq:mcemupdate} can be determined in a closed-form manner, as described below
 \begin{align}
    \label{eq:closeformupdate}
    \mu^*&=\sum_{i=1}^N \Big(R(\varGamma^i) \lambda^i\Big)\Big/\sum_{i=1}^N R(\varGamma^i)\\
    \sigma^*&=\sqrt{\sum_{i=1}^NR(\varGamma^i)(\lambda^i - \mu^*)^2\Big/\Delta}
 \end{align}
 where 
 \begin{align}
    \label{eq:closeformsupplement}
    \Delta=\bigg[\Big(\sum_{i=1}^N R(\varGamma^i)\Big)^2-\sum_{i=1}^N \Big(R(\varGamma^i)\Big)^2\bigg]\Big/\sum_{i=1}^N R(\varGamma^i).
 \end{align}

In this context, the reward of trajectories, denoted as $R(\varGamma^i)$, is considered as a form of weight, and it is inherently non-negative. These weights serve to gauge the significance of the corresponding sampled decision variables, $\lambda^i$. In accordance with equation \eqref{eq:closeformupdate}, decision variables associated with more substantial trajectory rewards are accorded higher weights during the update process for the policy's new parameters, $\mu^*$ and $\sigma^*$.
Once the reward values have converged, the iterative process concludes, and the algorithm yields the learned policy $\pi_{\theta^*}(\mu^*,\sigma^{*2})$. Subsequently, the decision variable $\lambda^*$ is sampled from the distribution $\pi_{\theta^*}(\mu^*,\sigma^{*2})$.

 \subsection{Deep policy learning}

The algorithm presented in Algorithm~\ref{alg:1} provides an effective approach to training policies for gate-traversing tasks. However, its multiple policy update iterations can lead to reduced time efficiency. To enhance the time efficiency of our proposed method, we employ two deep neural network policies designed to directly provide $\lambda$ and $t_p$ based on real-time observations during the gate-traversing process. These policies are trained using data collected through the repeated execution of Algorithm~\ref{alg:1}, which includes decision variables and corresponding high-dimensional observations, following a supervised learning paradigm \cite{highmpc2}. The complete training procedure is detailed in Algorithm~\ref{alg:2}.

\vspace{0.3cm}
\begin{algorithm2e}[H]
\SetAlgoNoLine %去掉竖线
\caption{\textbf{Learning neural network policies}}
\label{alg:2}
\KwIn{${\rm MLP}_\theta, {\rm MLP}_\rho, D_{\rm buf}=\{\}, \bm{x}_{\rm target}$}
\While{not tired}{
 \setlength{\parindent}{1em}
 Randomly initialize $\bm{x}_{t_0}^g, \bm{x}_{t_0}^q, \bm{x}_{},t=0$\\
    \While{flight not done}{
    \setlength{\parindent}{2em}
    $\lambda_t,t_{p}\leftarrow {\rm Algorithm\ } \ref{alg:1}(\bm{x}_{t_0}^q=\bm{x}_t^q,\bm{x}_{\rm target}=\bm{x}_{\rm target})$\\
		$D_{\rm buf}\leftarrow D_{\rm buf}\bigcup\{\bm{o}_t,\lambda_t,t_{p},t\}$\\
		$u_t={\rm MPC.solve}(\bm{x}_t^q,\bm{x}_t^g,\overline{\bm{x}}_{t_{p}}^g,\bm{x}_{\rm target},\lambda_t)$\\
		$\bm{x}_{t+d}^g\leftarrow f_{\rm gate}(\bm{x}_t^g),\bm{x}_{t+d}^q\leftarrow f_{\rm quad}(\bm{x}_t^q,\bm{u}_t)$\\ 
                }
		}
$\theta^*\leftarrow \arg\min_\theta\Vert {\rm MLP}_\theta(\bm{o}_t)-\lambda_t\Vert^2$\\
$\rho^*\leftarrow \arg\min_\rho\Vert {\rm MLP}_\rho(\bm{o}_t)-t_{p}\Vert^2$\\
\KwOut{Learned deep neural network ${\rm MLP}_{\theta^*},{\rm MLP}_{\rho^*}$}
\end{algorithm2e}
\vspace{0.3cm}

\begin{figure*}[t]
\centering
\subfigure[\quad hyMPC]{
\includegraphics[width=0.3\textwidth]{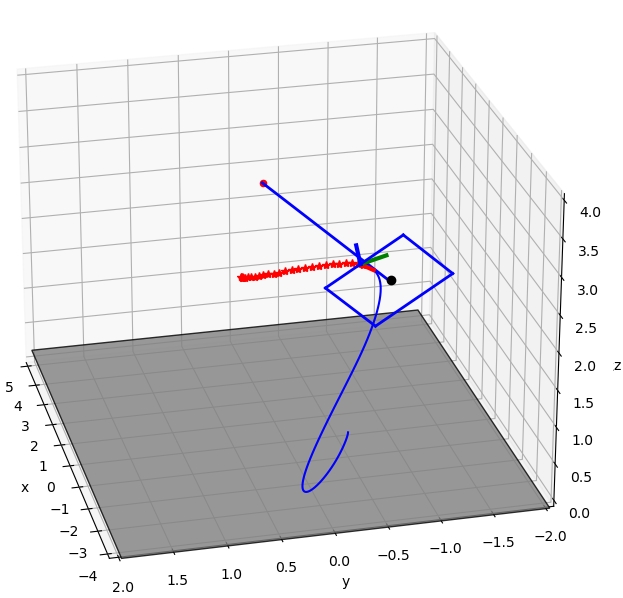}
}
\subfigure[\quad high-MPC]{
\includegraphics[width=0.3\textwidth]{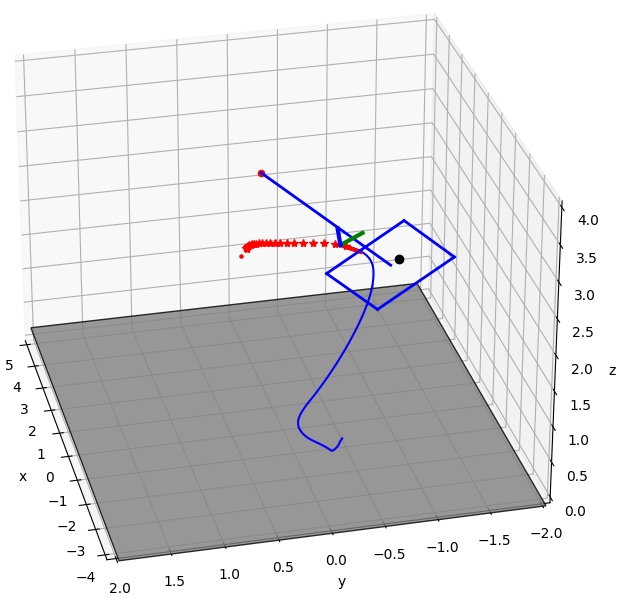}
}
\subfigure[\quad Standard MPC]{
\includegraphics[width=0.3\textwidth]{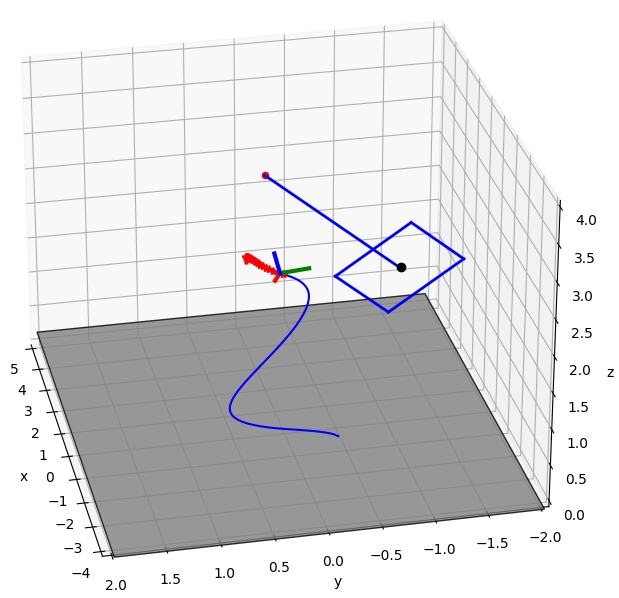}
}
\caption{A comparison between the proposed hyMPC, high-MPC, and standard MPC. (a) hyMPC with \emph{unknown} gate dynamics. (b) high-MPC with \emph{known} gate dynamics. (c) The standard MPC  with \emph{unknown} gate dynamics.}\label{fig:sucrate}
\end{figure*}

Contrary to Algorithm~\ref{alg:1}, in Algorithm~\ref{alg:2}, the traversal time interval is directly determined by a learned policy denoted as $\rm{MLP}_\rho$. This approach significantly reduces the computational burden associated with searching for the optimal traversal time interval. Initially, Algorithm 1 is executed iteratively, yielding a pair of parameters $\lambda_t$ and $t_p$ at each time step $t$. Subsequently, the MPC controller computes the predicted trajectory and control input $u_t$ for execution. Simultaneously, the parameters $\lambda_t$, $t_p$, and $t$, along with the relative state of the drone and the gate, denoted as $\bm{o}_t=\bm{x}_t^g-\bm{x}_t^q$, are stored in the replay buffer $D_{buf}$. Once the replay buffer $D_{buf}$ is filled, the data collection phase concludes. Two separate $\rm{MLP}$s are then trained using the data from $D_{buf}$. Specifically, $\rm{MLP}_\theta$ takes $\bm{o}_t$ as input and produces $\lambda_t$ as output, while $\rm{MLP}_\rho$ also uses $\bm{o}_t$ as input and generates $t_p$. Stochastic gradient descent is employed to train both neural networks, $\rm{MLP}_\theta$ and $\rm{MLP}_\rho$. The training objective is to minimize the mean-squared-error between the network outputs and the corresponding labels
\begin{equation}
\begin{aligned}
    \label{nnpolicytraining}
    \Vert{\rm MLP}_\theta(\bm{o}_t)&-\lambda_t\Vert^2\\
    \Vert{\rm MLP}_\rho(\bm{o}_t)&-t_p\Vert^{2}
\end{aligned}
\end{equation}
where the labels, $\lambda_t$ and $t_p$, are sampled from the replay buffer $D_{buf}$ along with their corresponding $\bm{o}_t$. Subsequently, Algorithm 2 concludes when the training error of the neural networks converges, resulting in the output of two neural networks denoted as ${\rm MLP}_{\theta^*}$ and ${\rm MLP}_{\rho^*}$. During the execution of gate-traversing tasks, at any given time $t$, ${\rm MLP}_{\theta^*}$ takes the current state of the drone and the gate, denoted as $\bm{o}_t$, as input and provides the policy parameter $\lambda_t$ for immediate use by the MPC controller. Similarly, ${\rm MLP}_{\rho^*}$ operates in the same manner as ${\rm MLP}_{\theta^*}$. Since the $\rm{MLP}$s calculate their output through a single forward propagation step, they offer greater convenience when employed during drone flight compared to directly using the proposed Gaussian policy. 

\begin{tablehere}
\renewcommand{\arraystretch}{1}
\tbl{Average traversal error and traversal time of the Gaussian policy, neural network policies, high-MPC \cite{highmpc2} and manual-MPC. \label{tab:table1}}
{
% \begin{adjustbox}{width=0.5\textwidth}
\begin{tabular}{ccc}
\toprule
Policy & Average error ($m$) & Average time ($s$)\\
\colrule
Gaussian & $0.124$ & $1.52$\\
%\hline
Neural network & $0.101$ & $1.49$\\
%\hline
high-MPC & $0.160$ & $1.18$\\
%\hline
manual-MPC & $0.121$ & $2.48$\\
\botrule
\end{tabular}
% \end{adjustbox}
}
\label{tab:singlesimulation}
\end{tablehere}

\section{Simulation Results}

\subsection{Simulation settings}

We have validated the efficacy of the proposed hyMPC through numerical simulations designed to control a drone as it traverses a swinging gate. 

The drone model employed in simulations is identical to that described in \cite{highmpc2}. Specifically, the length of the arm was $l=0.3m$, and the gravity acceleration was $g=9.81m/s^2$. The maximum angular velocity was set to $\omega_{max}=6.0rad/s$, and the thrust-to-weight ratio was set to $c\in [2.0, 20.0]$. 

For gate motion prediction and neural network policy learning, we employed a fully connected $\rm{MLP}$ with two hidden layers, each comprising $256$ units, and employed the exponential linear unit (ELU) as the activation function. The drone's starting podium was positioned at $[x^q,y^q,z^q,\dot{x}^q,\dot{y}^q,\dot{z}^q]=[-5,0,1.5,0,0,0]$, while the pivot point of the swinging gate remained fixed at $[x^p,y^p,z^p]=[2,0,3]$. The distance from the pivot point to the gate's center, denoted as $L$, measured $2$ meters (m). The gate's dimensions were specified with a height of $z_{h}=0.8m$ and a width of $y_{w}=1m$. Initial values for the gate's angle and angular velocity were randomly drawn from $[-\frac{\pi}{4},\frac{\pi}{4}]$ and $[-\frac{\pi}{20},\frac{\pi}{20}]$, respectively.

\begin{figurehere}
	\centering
	\includegraphics[width=2.5in]
	{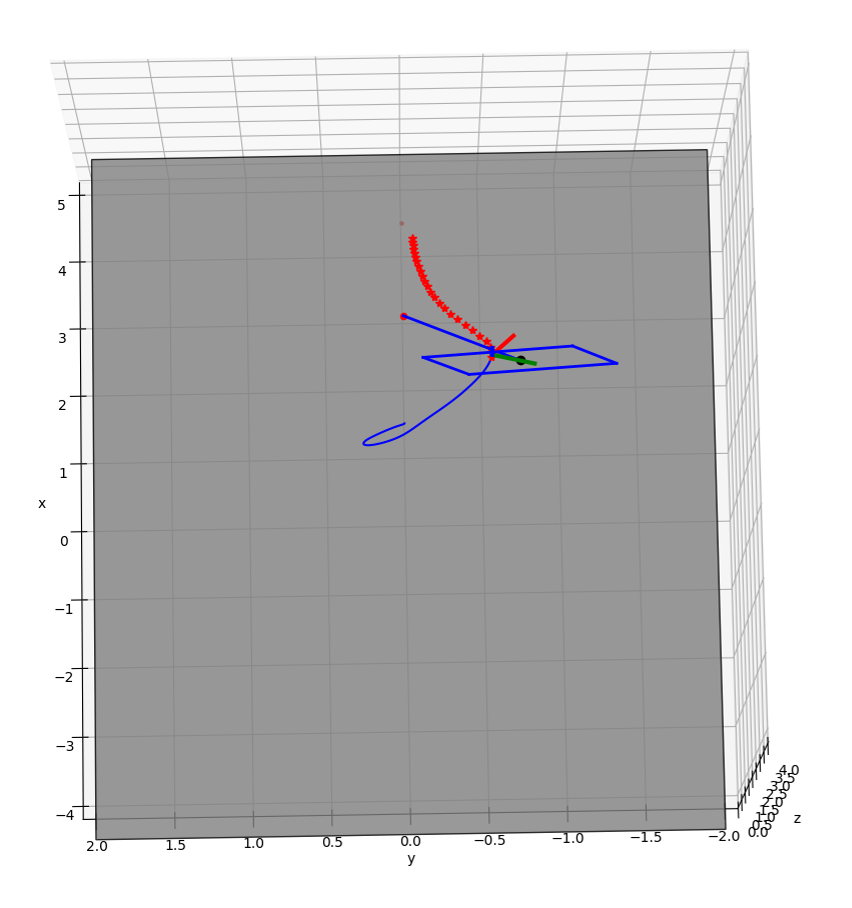}
	\caption{A flight in which the drone is initialized $1m$ away from the gate in the axial direction. At the start, the drone chose to turn left and backtrack to create sufficient axial space to facilitate the traversal task.}
	\label{fig:followfirst}
\end{figurehere}

We set the prediction time horizon for MPC, denoted as $t_{H}$, to 1 second (s), and the trustworthy prediction time horizon for the gate's motion prediction, represented as $\mathcal{T}_p$, was also set to $1s$. The discrete time interval $d$ was chosen as $0.02s$. Other parameters included $\eta=10$, $\zeta=3$, $\varepsilon=0.3m$, $\Omega=99,999$, and $\varrho=10$. The neural networks were implemented using PyTorch\cite{pytorch}, and the numerical simulation was conducted using CasADi\cite{casadi}.
For training Gaussian policies and deep policies, we utilized a laptop equipped with an \textit{Intel Core i7-11800h} CPU and an \textit{Nvidia Geforce RTX3060} graphics card. Durning the training of gaussian policies, the number of resampling in each step $N$ was $5$. The number of samples collected for the training of neural networks was $3000$. The neural networks took approximately $1$ hour to converge. 

\begin{figure*}[t]
\centering
\subfigure[]{
\includegraphics[width=0.23\textwidth]{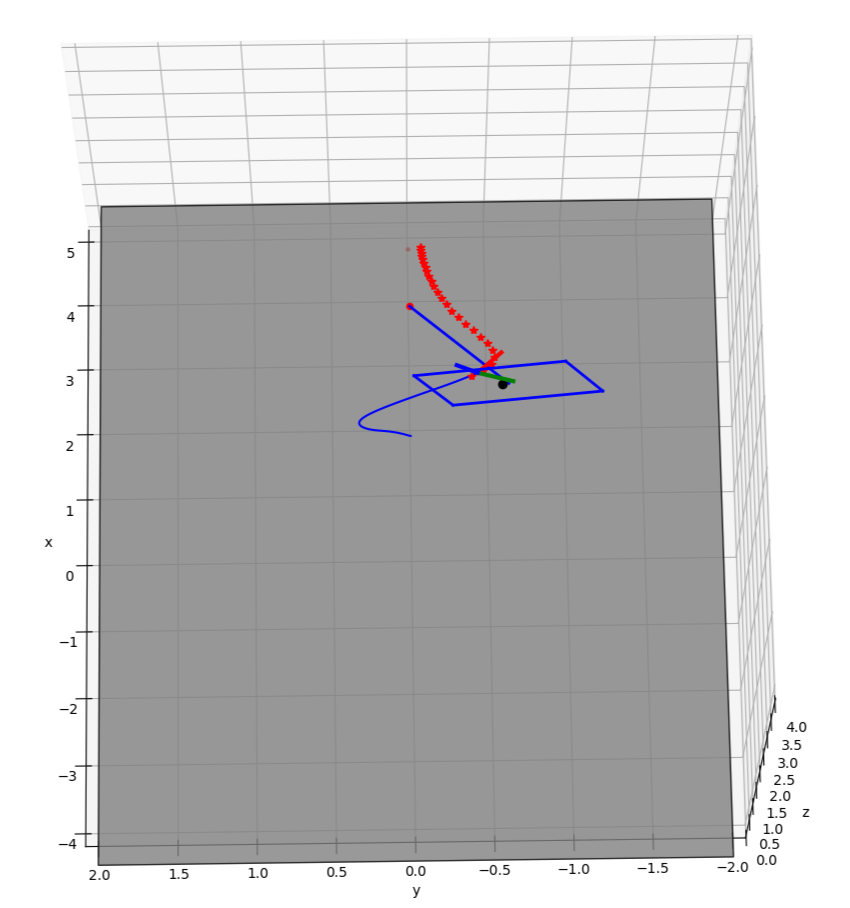}
\includegraphics[width=0.23\textwidth]{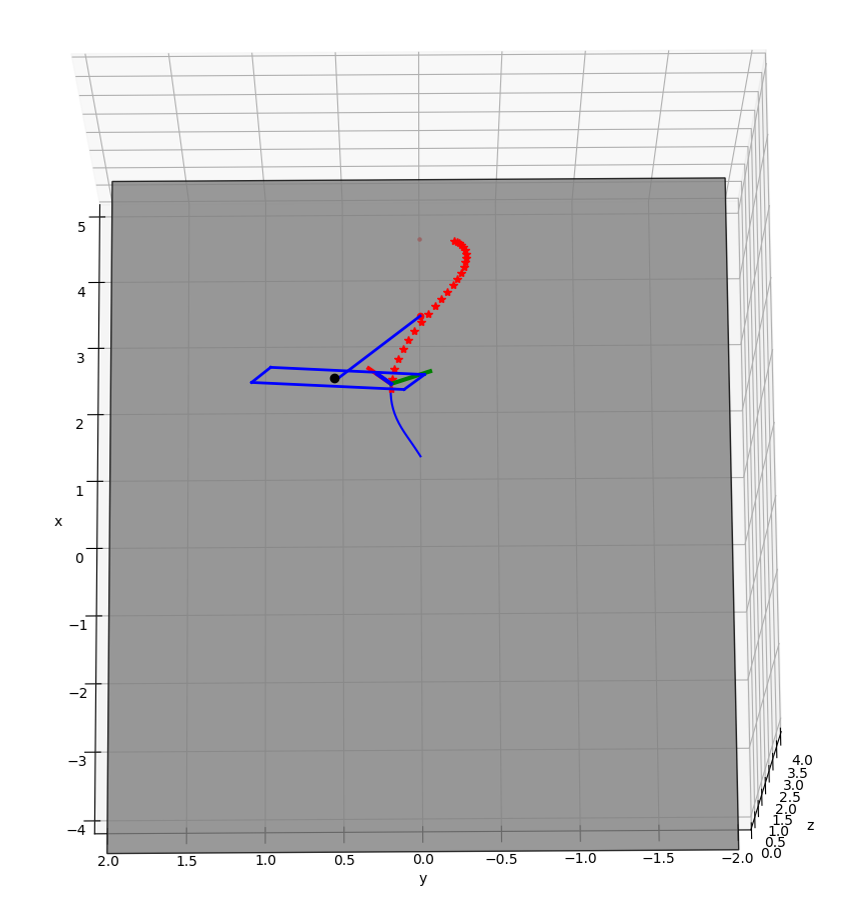}
}
\subfigure[]{
\includegraphics[width=0.23\textwidth]{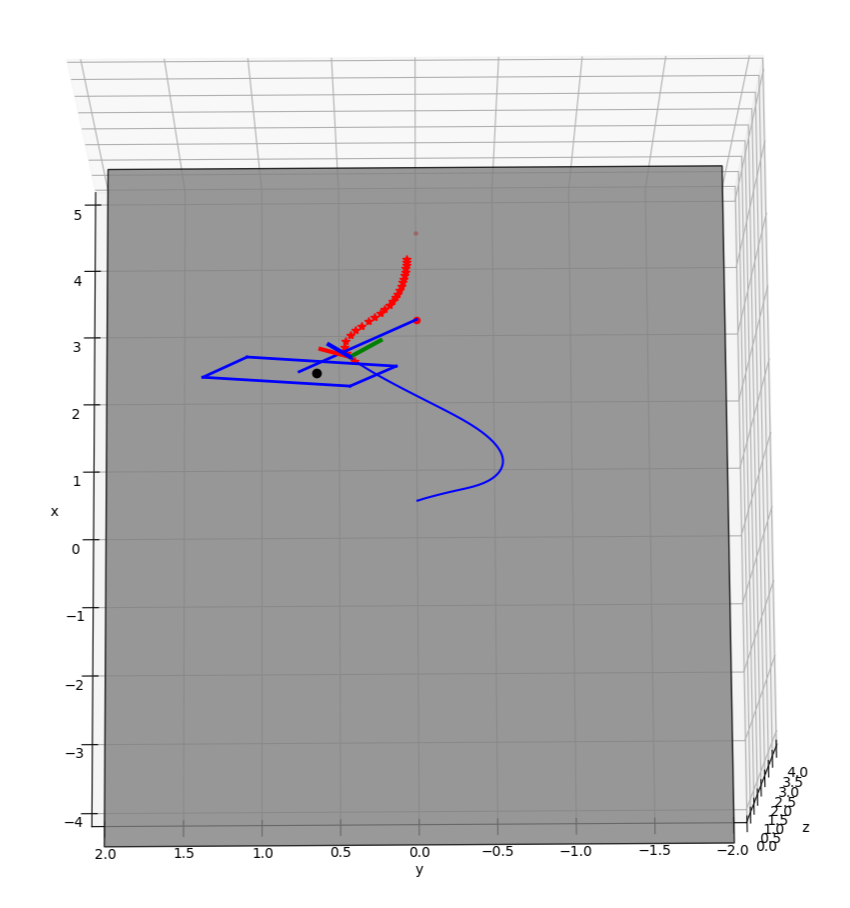}
\includegraphics[width=0.23\textwidth]{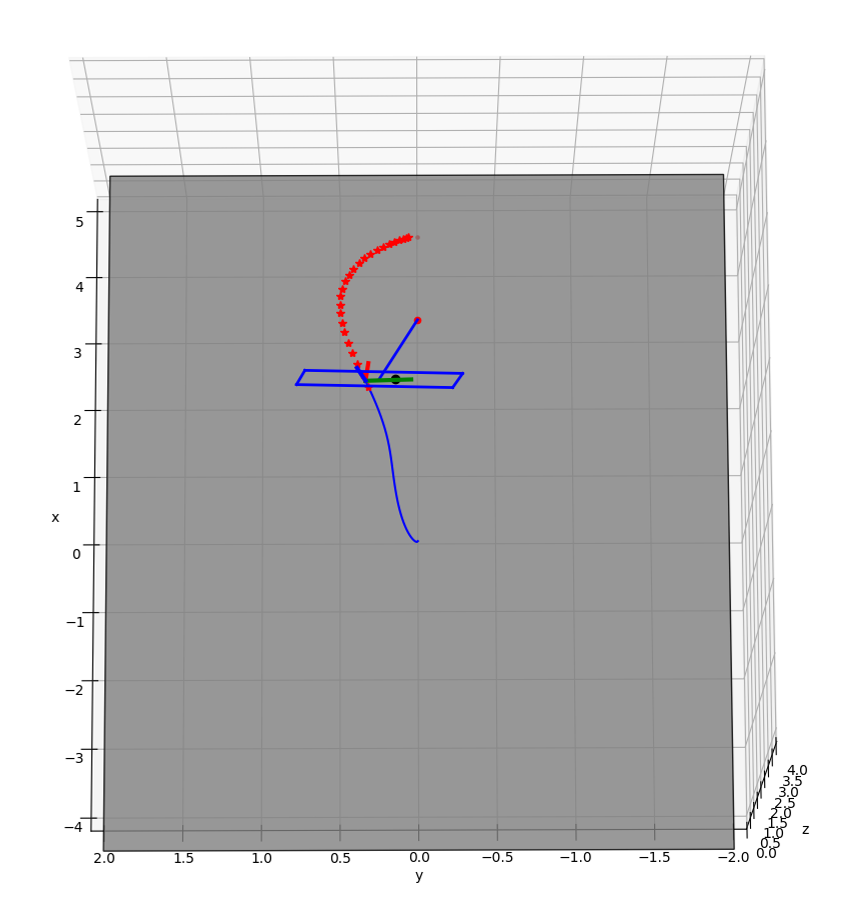}
}
\subfigure[]{
\includegraphics[width=0.23\textwidth]{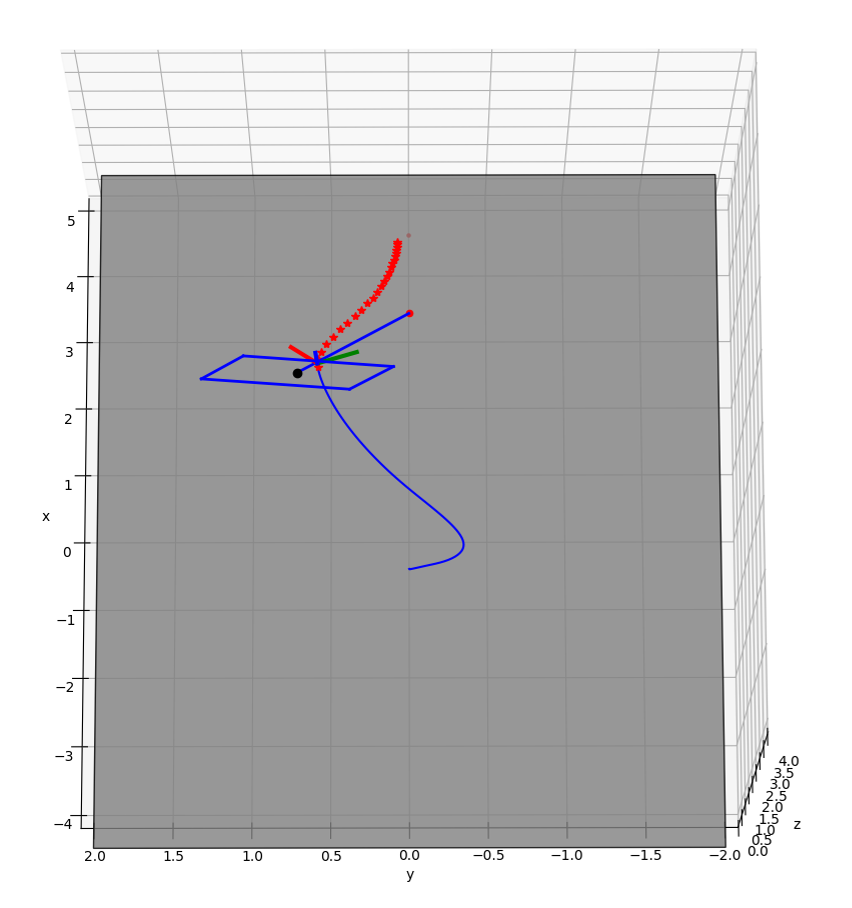}
\includegraphics[width=0.23\textwidth]{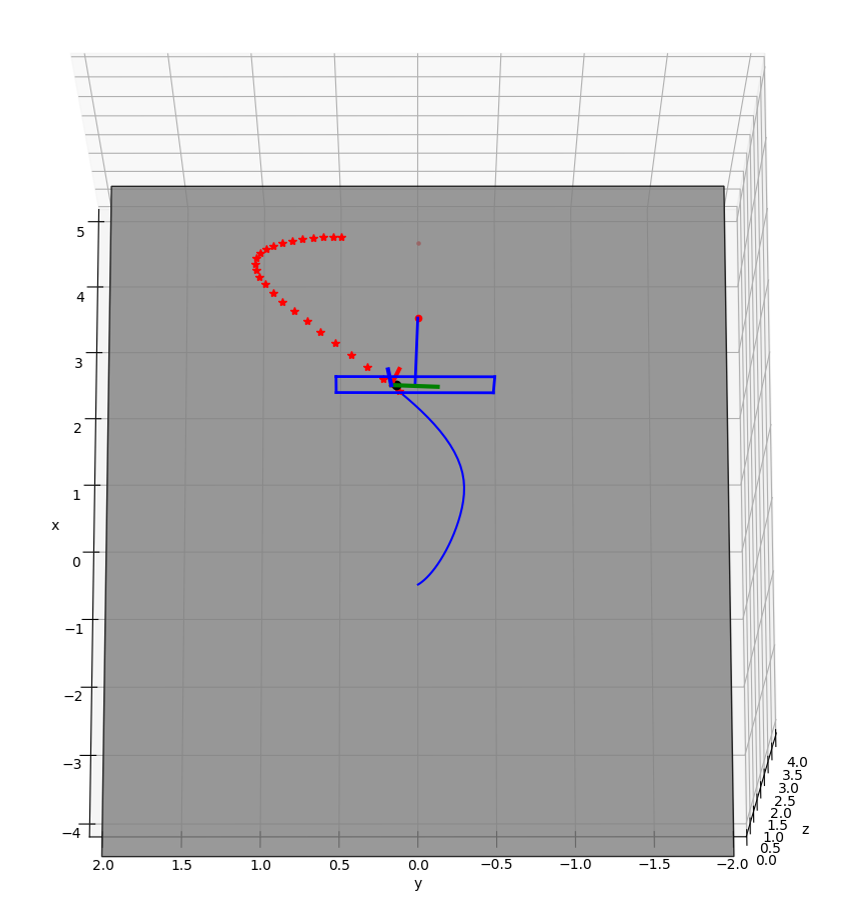}
}
\subfigure[]{
\includegraphics[width=0.23\textwidth]{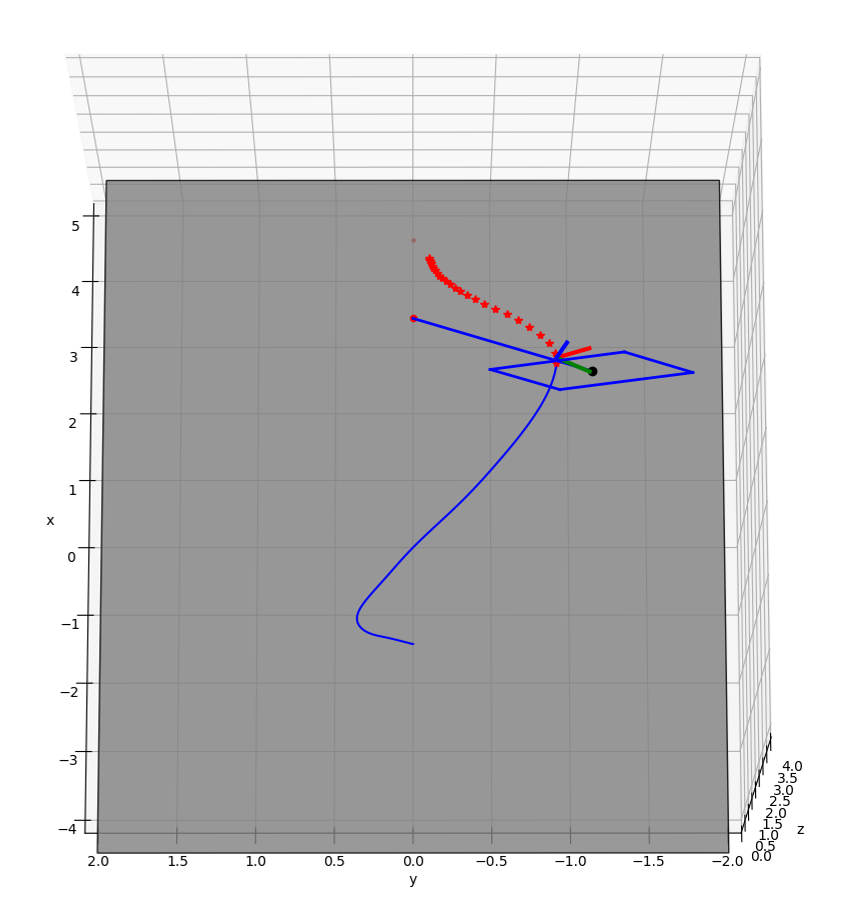}
\includegraphics[width=0.23\textwidth]{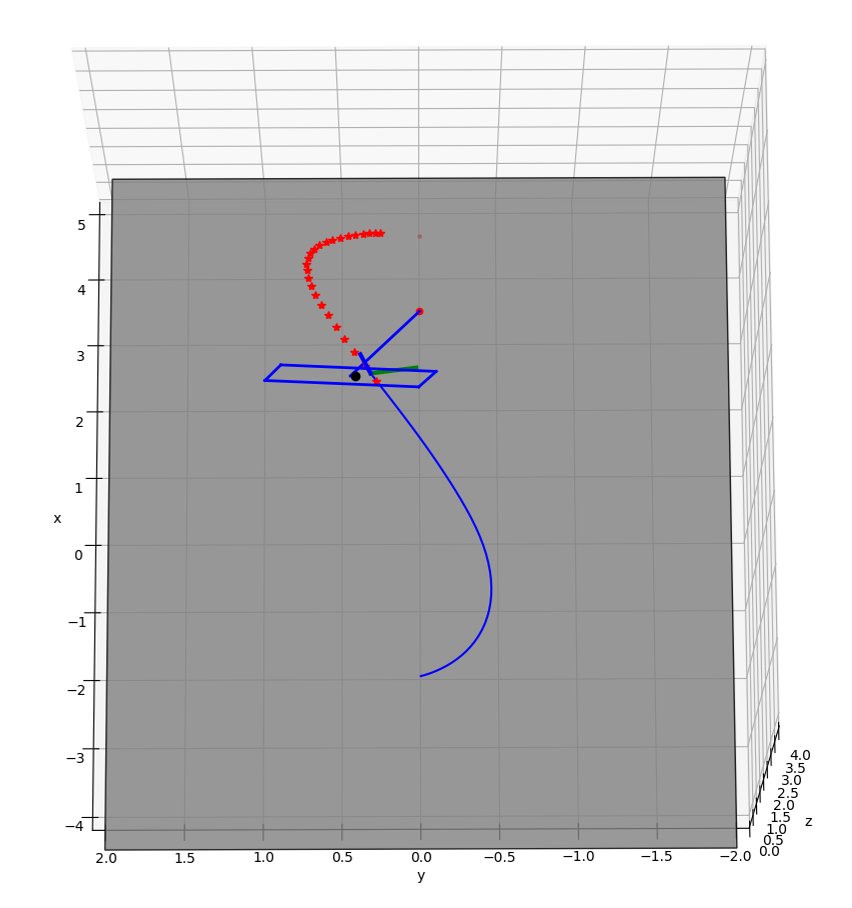}
}
\subfigure[]{
\includegraphics[width=0.23\textwidth]{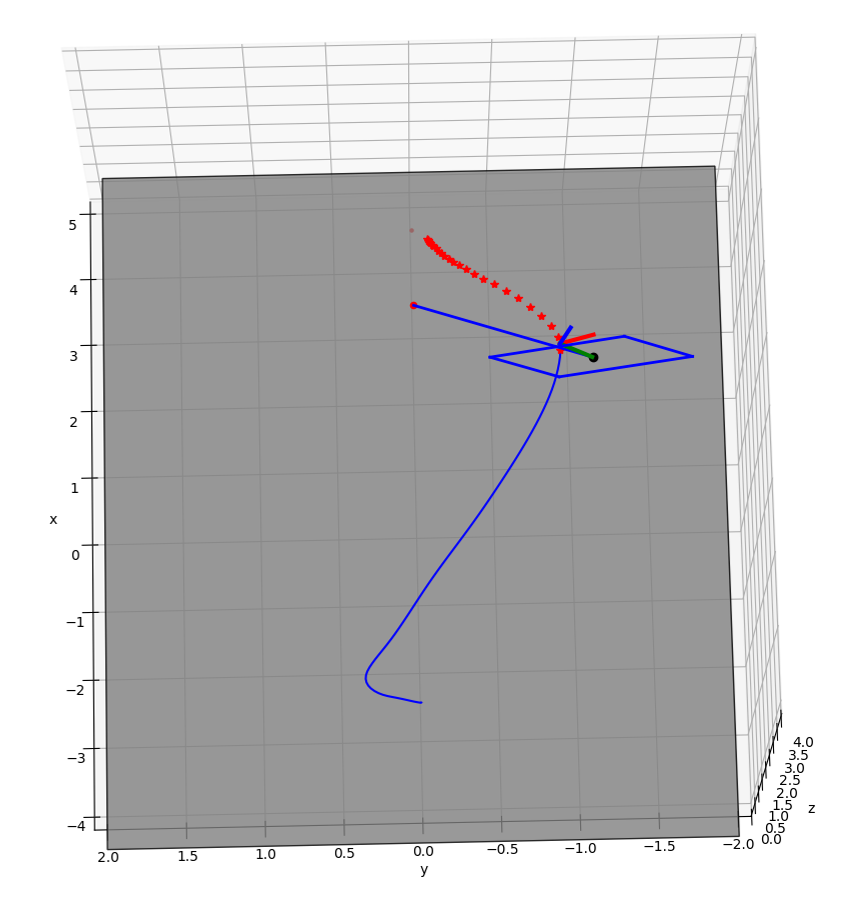}
\includegraphics[width=0.23\textwidth]{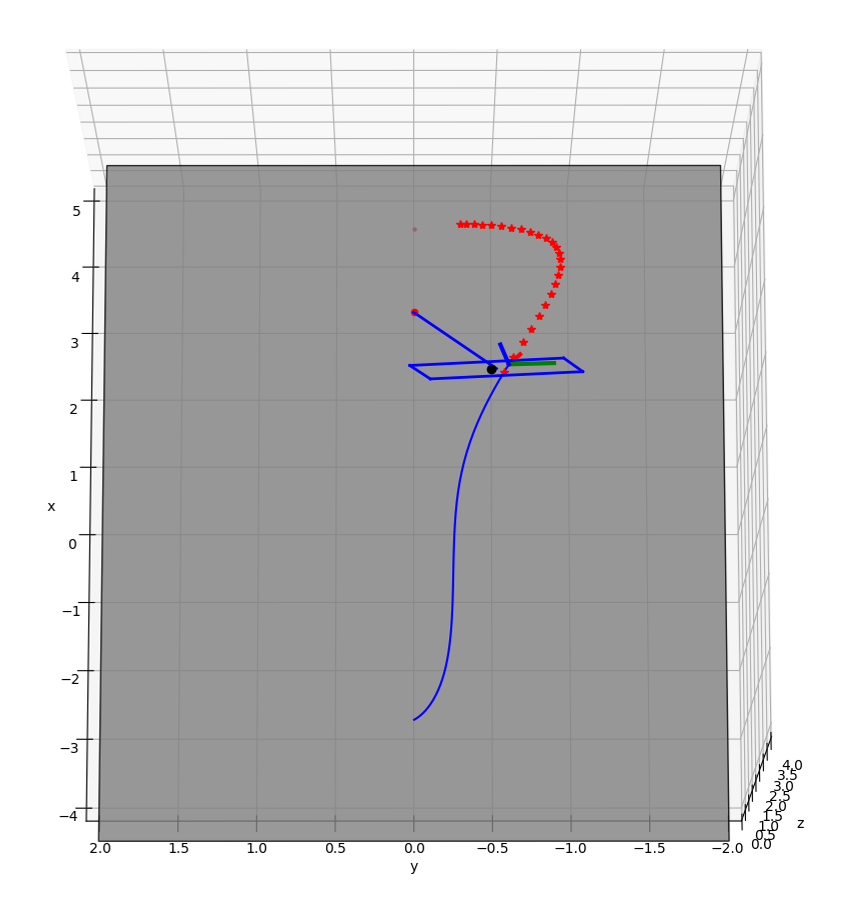}
}
\subfigure[]{
\includegraphics[width=0.23\textwidth]{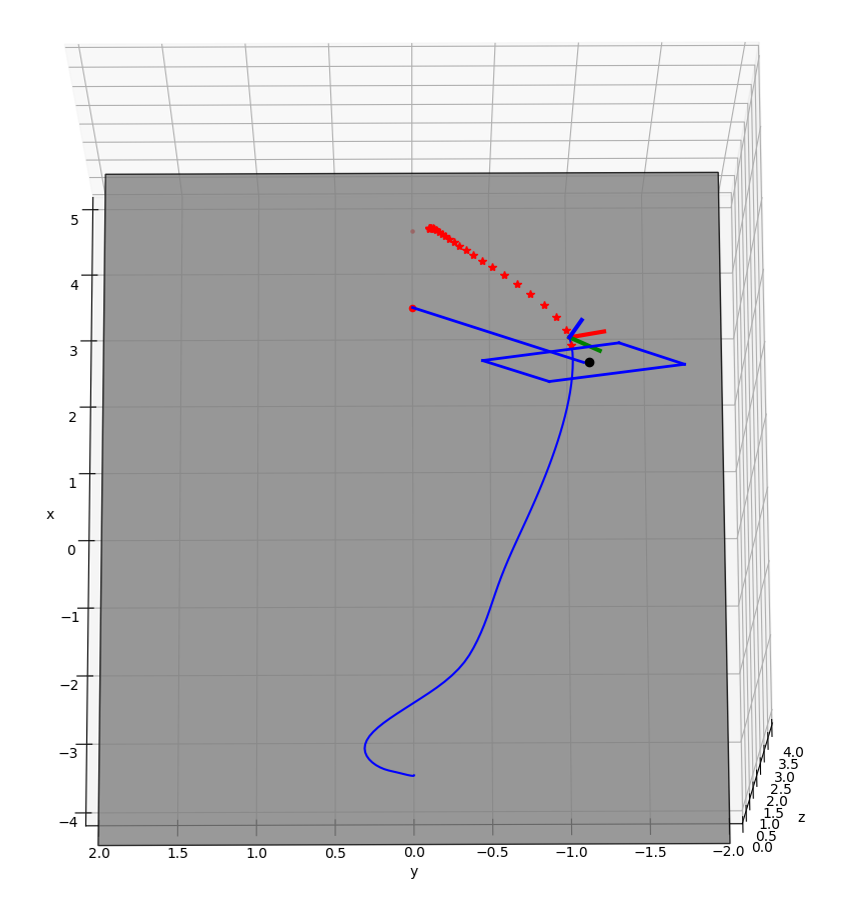}
\includegraphics[width=0.23\textwidth]{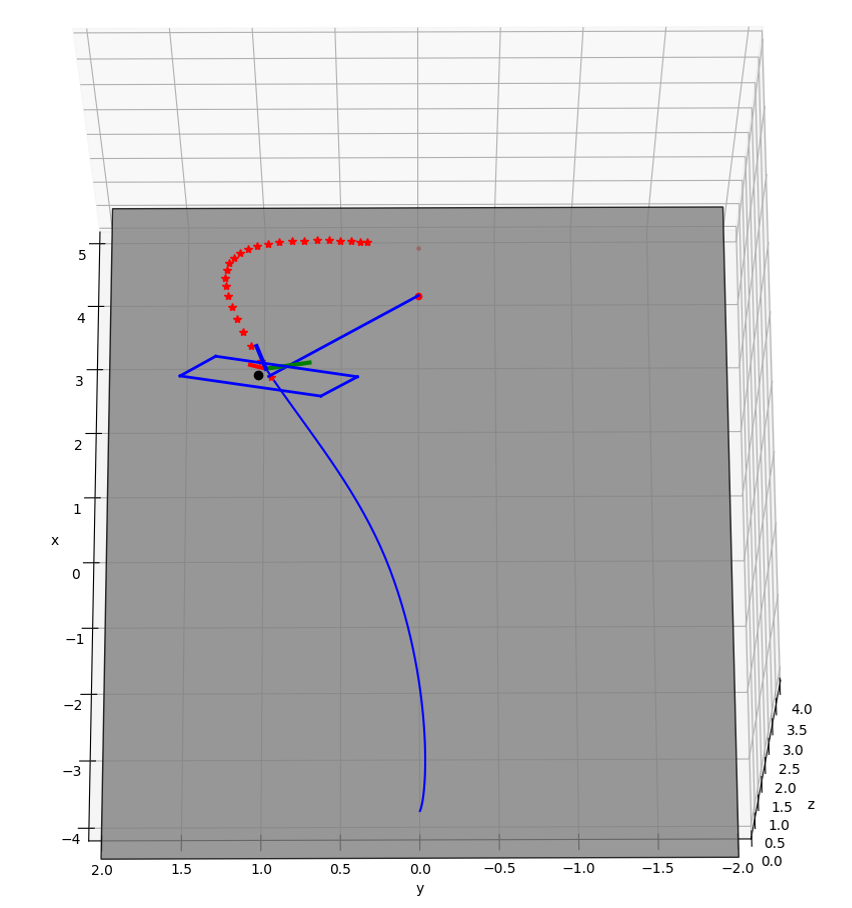}
}
\caption{A visualization of the comparison between the proposed framework hyMPC and high-MPC on different initial axial distances between the drone and the gate. In each subfigure, the left shows the flight produced by hyMPC, while the right displays the flight produced by high-MPC. (a) Initial distance of $1m$. (b) Initial distance of $2m$. (c) Initial distance of $3m$. (d) Initial distance of $4m$. (e) Initial distance of $5m$. (f) Initial distance of $6m$.}\label{fig:varydistance}
\end{figure*}

\begin{tablehere}
	\renewcommand{\arraystretch}{1}
	\tbl{Average traversal error and traversal time of the neural network policies in the multi-gate traversing task. \label{tab:table2}}
	{
		\begin{tabular}{cccc}
			\toprule
			Neural network & $1st$ gate & $2nd$ gate & $3rd$ gate\\
			\colrule
			Success rate ($\%$) & $100$ & $100$ & $100$\\
			Average error ($m$) & $0.097$ & $0.154$ & $0.167$\\
			%\hline
			Average time ($s$) & $1.506$ & $2.527$ & $3.865$\\
			\botrule
		\end{tabular}
		% \end{adjustbox}
}
\end{tablehere}

\subsection{Comparison} \label{section:comparison}

We conducted simulations to compare the performance of hyMPC with two baselines: high-MPC \cite{highmpc2} and standard MPC. Both high-MPC and standard MPC were configured with the same prediction time horizon $\mathcal{T}_p$ as hyMPC. High-MPC assumed known gate dynamics, while standard MPC utilized the same cost function as hyMPC but omitted the use of temporal spread weight. The experiments were executed $20$ times, each with randomly initialized gate dynamics.

\begin{table*}
\tbl{Average traversal error and traversal time of hyMPC and high-MPC in various initial distances. \label{tab:varydistance}}
{
% \begin{adjustbox}{width=0.8\textwidth}
% \normalsize
\begin{tabular}{cccccccc}
\toprule
\multirow{1}{*}{Method}        & $1m$ & $2m$ & $3m$ & $4m$ & $5m$ & $6m$ & Performance \\ \colrule
\multirow{2}{*}{high-MPC}        & $40$  & $60$ & $80$  & $90$ & $90$ & $95$ & Success rate (\%)  \\
                                 & $0.563$ & $0.367$ & $0.274$ & $0.253$ & $0.238$ & $0.176$ & Traversal error ($m$) \\ \midrule
\multirow{2}{*}{hyMPC} & $100$ & $95$ & $95$ & $95$ & $100$ & $100$ & Success rate (\%) \\
                                 & $0.156$ & $0.146$ & $0.141$ & $0.125$ & $0.101$ & $0.078$ & Traversal error ($m$)\\ \botrule
\end{tabular}
% \end{adjustbox}
}
\end{table*}

Considering the size of the drone, we set a lower bound of $0.4m$ for the position error to qualify as a successful gate pass. Fig. \ref{fig:sucrate} presents three trajectories generated by the three methods in a single experiment, serving as a demonstrative example. It illustrates that when confronted with extreme situations, such as the gate swinging to a large angle, standard MPC struggles to execute rapid maneuvers in order to successfully complete the traversal task. In our testing, the standard MPC achieved a success rate of approximately $50\%$ with an average traversal error of $0.410m$, attributed to its limited focus on the traversing time interval. In contrast, hyMPC achieved a $100\%$ success rate with an average traversal error of $0.101m$, indicating its capability to safely guide the drone through the gate. This can be attributed to the temporal spread weight employed in hyMPC, which directs the MPC's attention towards time intervals that are in closer proximity to the traversal time interval. 

To further evaluate the performance, we incorporated a relatively straightforward manual rule for determining the value of $\lambda$, aimed at contrasting and highlighting the advantages of the proposed reinforcement learning method. Specifically, the $\lambda$ was set to be: 
\begin{equation}
	\lambda_t=\frac{||x^{q}_{t}-x^{g}_{t}||}{||x^{q}_{t_0}-x^{g}_{t_0}||}
	\label{eq:lambda}
\end{equation}
where $\lambda$ was set to be smaller when the drone is closer to the gate, considering a smaller $\lambda$ indicates a greater tendency towards traversing the gate. The MPC working under this manual rule is named manual-MPC. We compared the traversal errors of the two proposed policies in hyMPC with high-MPC and manual-MPC. Table~\ref{tab:table2} presents the average traversal errors for the Gaussian policy, neural network policies, high-MPC and manual-MPC. 

The results clearly indicate that despite having a prediction horizon of $H=t_{H}/d=1/0.02=50$ steps for MPC, covering only a short segment of the entire flight trajectory, hyMPC yields more accurate flights than high-MPC. Remarkably, even in cases where the gate's dynamics remain unknown to hyMPC, they only require slightly longer traversal times compared to high-MPC. Moreover, in contrast to the hyMPC, the manual-MPC registers a significantly higher average traversal error and necessitates a considerably extended average traversal duration. This marked difference highlights the efficacy of the proposed learning-based approach in effectively identifying the subtle relationships between $\lambda$ and the dynamics of the flight's progression. 

We also conducted additional simulations of the gate traversal task, this time varying the initial distances between the drone and the gate for both high-MPC and hyMPC. Both methods were trained using the settings outlined in Section $5.1$. This comparison provided valuable insights, particularly in scenarios where the drone's initial radial distance to the gate is too close, yet it maintains some distance in the axial direction. In such cases, there is an increased risk of traversal task failure due to the inherent limitations of the drone's dynamics. Furthermore, when the drone is positioned at a considerable axial distance from the gate, the performance of MPC trajectory optimization can be compromised by the constrained prediction time horizon. The initial axial distance ranged from $1m$ to $5m$, and we conducted the simulation $20$ times at each distance, each time with randomly initialized gate dynamics. Fig. \ref{fig:varydistance} offers a visual representation of randomly sampled flights at different initial distances.

Table~\ref{tab:varydistance} presents the average success rates and traversal errors of hyMPC and high-MPC as we varied the axial distance between the drone and the gate. Notably, as the axial distance decreases, high-MPC exhibits a significant reduction in traversal errors, consistent with findings reported in \cite{highmpc2}. This observation suggests that high-MPC struggles to generalize across varying initial distances. In contrast, hyMPC maintains an exceptionally high success rate across different axial distances, even when the drone is initialized as close as $1m$ from the gate. This remarkable performance is attributed to our hybrid MPC optimization strategy, which makes intelligent decisions to temporarily follow the gate's motion when direct traversal is impractical. Fig. \ref{fig:followfirst} illustrates a typical flight showcasing this strategic approach. In this scenario, the drone was initialized just $1m$ away from the gate in the axial direction, while the gate was swinging to an extreme position. The drone opted to briefly reverse course and align with the gate in the radial direction before executing a successful traversal at an opportune moment. 

\subsection{Robustness test}
MPC methods are often challenged by model uncertainties and external disturbances. Specifically for drones, rapid battery depletion can lead to a decrease in the maximum current available to the motors, resulting in reduced thrust output. To address this, we designed a robustness test focusing on thrust degradation. In the MPC planning phase, the drone's maximum thrust, $c_{max}$, was maintained at the predefined value of $20$. However, during the state updating phase, the drone's maximum thrust, $\tilde{c}_{max}$, was deliberately reduced to various extents to simulate thrust degradation. The outcomes of this simulation are depicted in Fig. \ref{fig:robustness}. 

Considering the existence of gravity $g=9.81m/s^2$, the drone's thrust-to-weight ratio should at least be larger than $10.0$. Thus, the robustness test started from $\tilde{c}_{max} = 0.6c_{max} = 12.0$. The hyMPC, as proposed, exhibited significant robustness in the face of varying degrees of thrust attenuation. With the increase in thrust reduction, there was a corresponding extension in the time required for traversal. This effect can be ascribed to the reduced maneuverability of the drone when functioning under a diminished maximum thrust. Importantly, when the reduction in maneuverability was kept within a specific threshold, the average traversal error stayed within acceptable boundaries. This suggests that the hyMPC was capable of effectively carrying out traversal tasks even under compromised thrust conditions. 

\begin{figurehere}
	\centering
	\includegraphics[width=3.2in]
	{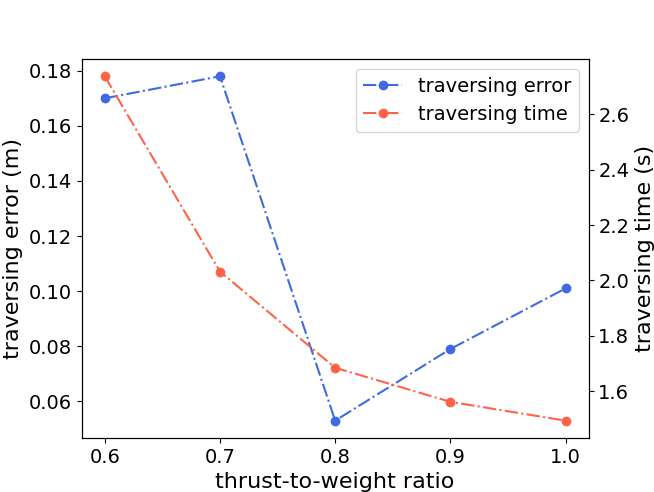}
	\caption{Average traversing errors and traversing times of hyMPC under different degrees of thrust degradation. The other settings were the same as the test demonstrated by Section \ref{section:comparison}. }
	\label{fig:robustness}
\end{figurehere}

\subsection{Multi-gate traversing}

Having successfully controlled the drone to traverse a single swinging gate with unknown dynamics, we extended our task to involve the continuous traversal of multiple moving gates, aligning more closely with real-world drone applications. We configured this task with three swinging gates with the same dynamic parameters as the comparison held in Section $5.2$, each separated by a spacing of $4m$. The experiment was conducted $20$ times, initializing the gate dynamics randomly for each run. Throughout the flight, the three gates swing independently, and the drone is required to traverse them seamlessly without interruption.

In this task, the drone consistently takes the state of the next gate to be traversed as the input $\bm{x}_{way}$, and designates the waypoint $2m$ behind the next gate as the input $\bm{x}_{target}$ for the proposed neural network policies in hyMPC. Fig. \ref{fig:multigate} depicts one of the $20$ trajectories generated during the experiment. Table \ref{tab:table2} presents the average traversing error and the average traversing time for all three gates across the $20$ trajectories. The proposed neural network policies achieved a $100\%$ success rate in traversing all the three gates. The multi-gate traversing task serves as a compelling demonstration of the adaptability of hyMPC to novel tasks that differ from those encountered during the training phase. This showcases the policies' versatility and their capacity to handle real-world drone scenarios beyond their initial training scope. 

These results underscore the capability of hyMPC to effectively control the drone for continuous traversal of multiple swinging gates with unknown dynamics. 

\begin{figurehere}
	\centering
	\includegraphics[width=3in]
	{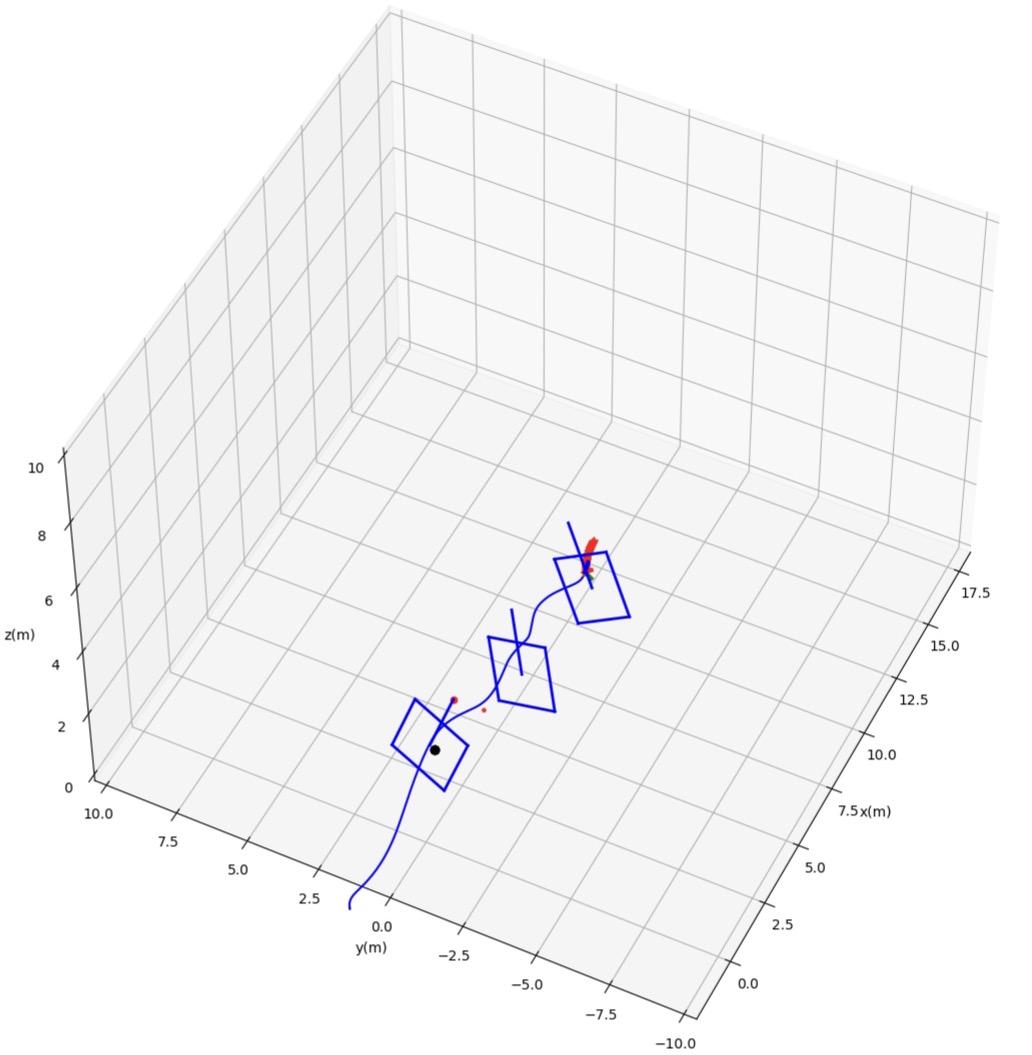}
	\caption{A drone traversing three swinging gates continuously guided by the proposed neural network policies in hyMPC.}
	\label{fig:multigate}
\end{figurehere}

\section{Conclusions}

In this study, we have introduced a novel framework that combines real-time MPC and offline RL techniques to navigate drones through swinging gates with unknown dynamics. Central to this approach is the on-the-fly training of an MLP to predict gate motions, facilitating high-level trajectory planning. The traversal task is thoughtfully divided into two subtasks, namely gate-following and gate-traversing, each guided by meticulously designed cost functions within MPC. A Gaussian policy is learned at the high-level to optimally blend these subtasks. Additionally, we have enhanced the time efficiency of our method by incorporating a policy search algorithm and supervised learning to train deep neural network policies that furnish high-level decision variables.
Our simulations have demonstrated the efficacy of our proposed hyMPC in comparison to two established alternatives: high-MPC and standard MPC. The proposed hyMPC consistently outperforms both in terms of achieving more accurate flights while requiring less prior knowledge. Furthermore, our approach exhibits superior generalization capabilities, particularly evident across varying initial axial distances between the drone and the gate. Given the common scenario of unknown environmental dynamics and the potential for drones to initiate tasks at various distances from obstacles or destinations in real-world applications, our approach represents a highly practical choice for deployment on real drones.

Despite its successes, the proposed hyMPC does have some limitations. One area for improvement is the efficiency of the policy update process, especially during Gaussian policy learning, where multiple MPC optimizations are necessary to select the traversal time within the trustworthy prediction horizon. This data-collection loop can be time-consuming, particularly for large-scale flight tasks that span extended periods. In future work, we intend to address this issue and enhance the overall time efficiency of the policy search method.
Additionally, we plan to extend hyMPC to handle drone traversing tasks that involve multiple gates with various dynamics. This extension will allow us to explore the robustness and adaptability of our approach in more complex scenarios.

\bibliographystyle{ws-us}
\bibliography{fzh}%self defined template

\end{multicols}
\end{document}